\documentclass[conference]{IEEEtran}
\usepackage[T1]{fontenc}
\usepackage[cmex10]{amsmath}
\usepackage{array}
\usepackage{booktabs}
\usepackage{siunitx}
\usepackage{amsfonts}
\usepackage[left=0.77in, right=0.77in, top=1in, bottom=0.78in]{geometry}
\usepackage{algorithm}
\usepackage[noend]{algpseudocode}

\usepackage{adjustbox}
\usepackage{multirow}
\usepackage{xcolor}
\usepackage{cite}
\usepackage{mathtools}

\usepackage[symbol]{footmisc}
\def\BibTeX{{\rm B\kern-.05em{\sc i\kern-.025em b}\kern-.08em
    T\kern-.1667em\lower.7ex\hbox{E}\kern-.125emX}}
\usepackage{graphicx}
\usepackage{pstricks}
\usepackage[utf8]{inputenc}
\usepackage{amssymb}
\usepackage{amsthm}
\usepackage{relsize}
\usepackage{bbm}
\usepackage{bm}
\usepackage[font={small}]{caption}
\usepackage{comment,color,soul}
\usepackage{textcomp}
\makeatletter
\def\thm@space@setup{\thm@preskip=2pt
        \thm@postskip=2pt \itshape}
\makeatother
\newtheoremstyle{newstyle}
{} 
{} 
{\mdseries} 
{} 
{\bfseries} 
{.} 
{ } 
{} 
\usepackage{url}
\theoremstyle{newstyle}
\usepackage{subcaption}

\theoremstyle{definition}

\theoremstyle{remark}

\IEEEoverridecommandlockouts

\begin{document}
\title{A Pre-defined Sparse Kernel Based Convolution for Deep CNNs}
\author{\IEEEauthorblockN{Souvik Kundu,$^{\ddagger}$ Saurav Prakash,$^{\ddagger}$ Haleh Akrami, Peter A.~Beerel, Keith M.~Chugg}
\thanks{$^{\ddagger}$Authors have equal contribution.}
\thanks{This work was partly supported by NSF funding, including grant \#1763747.} 
\IEEEauthorblockA{\textit{Ming Hsieh Department of Electrical and Computer Engineering} \\
\textit{University of Southern California}\\
Los Angeles, California 90089, USA \\
\{souvikku, sauravpr, akrami, pabeerel, chugg\}@usc.edu}
}
\maketitle

\begin{abstract} 
The high demand for computational and storage resources severely impedes the deployment of deep convolutional neural networks (CNNs) in limited resource devices. Recent CNN architectures have proposed reduced complexity versions (e.g,. SuffleNet and MobileNet) but at the cost of modest decreases in accuracy. This paper proposes pSConv, a pre-defined sparse 2D kernel based convolution, which promises significant improvements in the trade-off between complexity and accuracy for both CNN training and inference.  To explore the potential of this approach, we have experimented  with two widely accepted datasets, CIFAR-10 and Tiny ImageNet, in sparse variants of 
both the ResNet18 and VGG16 architectures. 
Our approach shows a parameter count  reduction of up to $4.24 \times$ with modest degradation in classification accuracy relative to that of standard CNNs. Our approach outperforms a popular variant of ShuffleNet using a variant of ResNet18 with pSConv having $3 \times 3$ kernels with only four of nine elements not fixed at zero. In particular, the parameter count is reduced by $1.7 \times$ for CIFAR-10 and $2.29 \times$ for Tiny ImageNet with an increased accuracy of $\sim4\%$.
\end{abstract}

\IEEEpeerreviewmaketitle

\section{Introduction and Motivation}
\label{sec:intro}
Convolutional Neural Networks (CNNs) play a significant role in driving a variety of technologies including computer vision \cite{he2016deep, girshick2014rich, krizhevsky2012imagenet}, natural language processing (NLP) \cite{young2018recent}, and speech recognition \cite{abdel2014convolutional}. In an effort to improve classification performance, CNN sizes have grown dramatically from 1 million (1M) in 1998  \cite{lecun1998mnist}, to 60M in 2012 \cite{krizhevsky2012imagenet}, to more recent networks with 10 billion trainable parameters \cite{coates2013deep}. Complexity has become an important consideration as CNNs have come into use in practical systems.  In particular, the storage and computational complexities, as well as energy consumption, are important considerations in both training and inference modes.  As the size of datasets increases and the range of inference tasks broadens, training complexity will continue to be an important consideration.  Similarly, many emerging applications require inference processing of trained networks on edge computational platforms that are severely constrained in terms of storage, computation, and energy consumption \cite{hegde2018ucnn}.  

The methods proposed to address the issue of large models can be categorized into three general areas.  One general approach is  \emph{pre-defined constrained filter design}, wherein the standard CNN filter kernels are constrained in some fashion to reduce the computational and storage complexity.  Primary examples in the category are MobileNet and ShuffleNet which rely on separable and grouped filter kernels, respectively.  
A second group of techniques is \emph{pruning during training}, wherein parameters are removed from the model as training is performed to produce a low-complexity trained model for inference.  
A third category for complexity reduction is \emph{weight quantization}, wherein the trained weights are grouped and/or quantized to save storage, possibly including an iterative training procedure \cite{courbariaux2015binaryconnect, zhou2017incremental, rastegari2016xnor}.  

An important distinction between the \emph{pre-defined constrained filter} approach and the \emph{pruning during training} approach is that only the former can be used to significantly reduce the complexity of training. Furthermore, architectural implementation can be considered in pre-defined constrained filters.  
For example,  structure can be imposed so that storage and retrieval of the weights is simplified in software or hardware implementations. 
Pruning, in contrast, typically results in patterns of connectivity that are not structured or predictable \cite{han2015learning, guo2016dynamic, mao2017exploring}.  
Moreover, pruning has been shown to be more effective in the fully-connected layers than in the convolutional layers \cite{wen2016learning} whereas the pre-defined constrained filter design approach directly implements complexity reduction in the convolutional layers.
Weight quantization is a more general optimization method that can be combined with the concept of pruning as well as pre-defined constrained filter design.  In all cases of parameter reduction, one is typically concerned with understanding the trade-off between  complexity reduction and inference performance.

\begin{figure*}[h]
\includegraphics[width=0.8\linewidth]{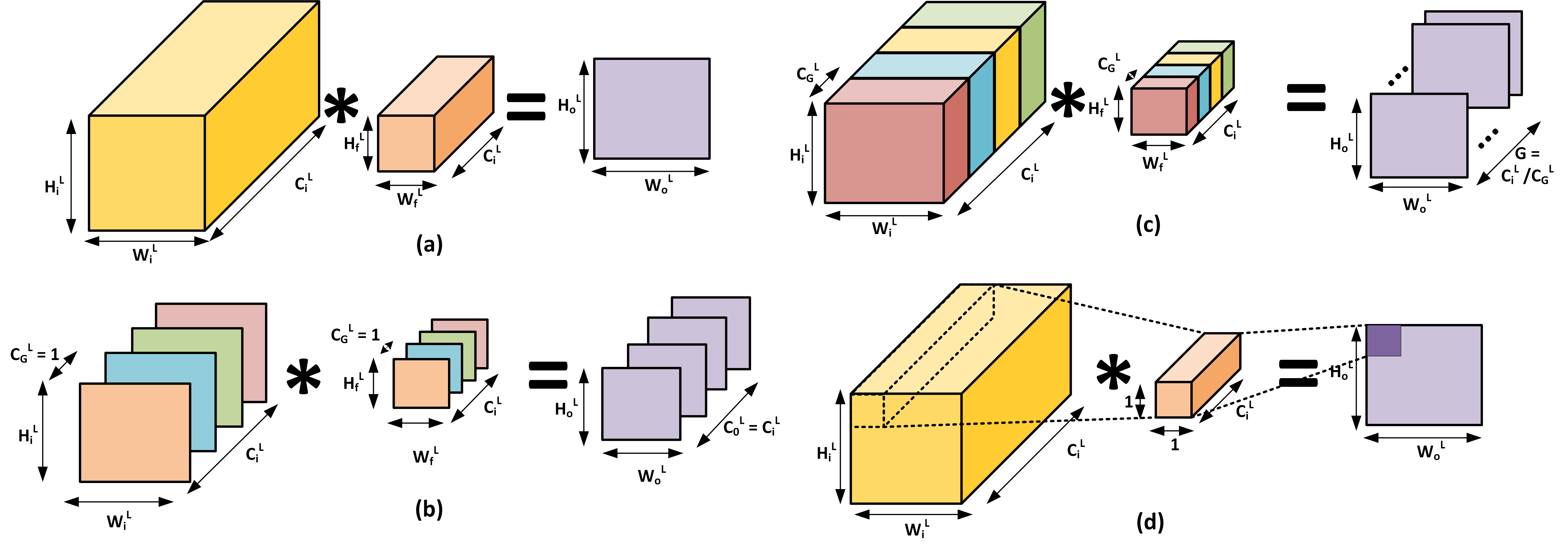}
\centering
\caption{Different types of convolutions (a) standard full  channel convolution (SFCC), (b) depth-wise convolution (DWC), (c) group-wise convolution (GWC), (d) point-wise convolution (PWC).}
\label{fig:diff_CNNs}
\end{figure*}

In this paper, we propose a new method of model complexity reduction which is in the category of pre-defined constrained filter design approaches -- i.e., pre-defined Sparse Convolutional (pSConv) layers.  The approach is to pre-define sparse patterns in the 2D filter kernels.  For example, if a  $3 \times 3 \times 20$ filter is considered in a standard CNN layer (i.e., $3 \times 3$ 2D kernels combined across 20 input channels), we would set, for example, 5 of the 9 coefficients in each channel plane to be zero.  These zero locations are defined before training\footnote{The precise pattern used is chosen pseudo-randomly in this work.} and are held fixed during the training and inference processes.  
This work extends our previous investigation into using pre-defined sparsity in fully-connected layers \cite{dey2019pre} and demonstrates similar benefits for CNN layers.  While efficacy in fully-connected layers motivated this work, it is not obvious that pre-defined sparsity would be effective in CNNs because CNNs, by definition, are sparse in a spatial sense.  In this paper, we demonstrate the effectiveness of pSConv layers in variants of ResNet \cite{he2016deep} and VGG \cite{simonyan2014very} on the CIFAR-10 \cite{krizhevsky2009learning} and Tiny ImageNet \cite{le2015tiny} datasets.  Most notably, we have reproduced ShuffleNet \cite{zhang2018shufflenet} for ResNet  experiments and the proposed pSConv CNNs consistently outperform ShuffleNet in terms of the complexity-performance trade-off.  Specifically, pSConv networks with complexity comparable to ShuffleNet typically outperform ShuffleNet by 4-5\% in absolute accuracy or 6-7\% in relative accuracy.  ShuffleNet has been shown to outperform MobileNet \cite{howard2017mobilenets} under similar trade-off metrics which implies that the pre-defined sparse convolutional approach is an attractive state of the art model complexity reduction method for both training and inference.
Furthermore, we observe complexity reductions of up to 70\% with negligible degradation in accuracy relative to standard convolution based CNNs.  

The remainder of the paper is organized as follows.  In Section \ref{sec:rel_work}, we describe pre-defined constrained filter design methods proposed in the literature and describe how MobileNet and ShuffleNet are designed using these tools.  Section \ref{sec:spCNN} describes our proposed pSConv approach and includes a discussion of considerations for realizing the complexity reduction advantages in practice as well as expressions characterizing the computational complexity (in floating point operations (FLOPs)) for ShuffleNet and pSConv networks. 
 Experimental results are presented in Section \ref{Experment_eval} while conclusions and suggestions for further research are summarized in Section \ref{sec:concl}.

\section{Related Work and Background}
\label{sec:rel_work}


In subsection A of this section, we describe various constraints on convolutional filters and in subsection B we describe how these approaches are combined to produce various efficient network architectures, including MobileNet and ShuffleNet.  
\begin{figure}[h]
\includegraphics[width=0.7\linewidth]{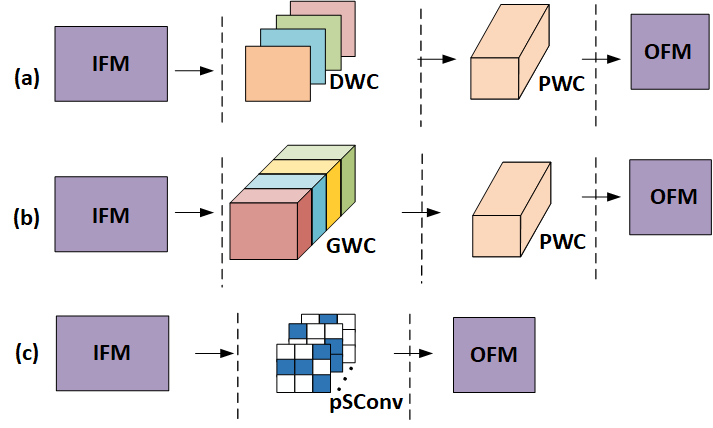}
\centering
\caption{Latency comparison between different types of convolutions.}
\label{fig:latency_0_1}
\end{figure}

\subsection{Pre-defined Constrained Filters}

Convolutional filters (CONVs) of CNN architecture can be broadly classified into four categories based on  constraints as shown in Fig.~\ref{fig:diff_CNNs}: (a) standard full channel (SFCC) \cite{lecun1999object}, (b) depth-wise (DWC) \cite{vanhoucke2014learning}, (c) group-wise (GWC) \cite{krizhevsky2012imagenet}, and (d) point-wise (PWC) \cite{szegedy2015going} convolutions. 
In all cases illustrated, the input feature map (IFM) dimension is assumed to be 
$H_i^L \times W_i^L \times C_i^L$, where $C_i^L$ is the channel depth. 
In Fig. \ref{fig:diff_CNNs}(a), a SFCC filter of dimension $H_f^L \times W_f^L \times C_i^L$ produces a single channel of output feature map (OFM) of size $H_o^L \times W_o^L$, after being convolved with the IFM. To obtain multiple channels in the output feature map, multiple of these filters are used.  
For the DWC shown in Fig. \ref{fig:diff_CNNs}(b), each 2D  kernel ($H_f^L \times W_f^L$)  is convolved with a single channel of the IFM to produce the corresponding OFM; thus $C_i^L$ 2D kernels will produce an OFM of dimension $H_o^L \times W_o^L \times C_i^L$.  This requires $C_i^L$ times less computations as compared to SFCC, but the output features capture no information {\em across} channels. 

Group-wise convolution, which provides a compromise between SFCC and DWC,  is shown in Fig. \ref{fig:diff_CNNs}(c). 
A single channel of the OFM is computed by convolving a group of channels from the IFM with  $C_G^L$ channels, with a CONV filter of size $H_f^L \times W_f^L \times C_G^L$.   Thus, with a total number of groups $G = (C_i^L/C_G^L)$, the same CONV filter of dimension $H_f^L \times W_f^L \times C_i^L$ provides an OFM of size $H_o^L \times W_o^L \times (C_i^L/C_G^L)$. 
With a channel per group of one, the GWC based approach reduces to DWC, while GWC with a single group is  equivalent to  SFCC.
Typically, the number of groups $G$ is within the set $\{2, 4, 8, 16\}$ and the best choice is network architecture dependent \cite{ioannou2017deep}.  In fact, in many cases, the optimal number depends on the location of the filter within the network \cite{ioannou2017deep}.

Finally, Fig. \ref{fig:diff_CNNs}(d) illustrates PWC in which the 2D kernel dimension is $1 \times 1$, thus generating a single OFM channel using far less computation. For example, compared to a $3 \times 3$ 2D kernel dimension, the PWC has $9\times$ less computational complexity. However, output feature points generated through this approach do not contain any embedded information {\em within} a channel.

\subsection{Efficient Network Architectures}



Many well-known network architectures use a combination of the approaches shown in Fig. \ref{fig:diff_CNNs} that have been 
empirically evaluated and optimized through numerous experiments.
A combination of GWC and PWC was used in \cite{ioannou2017deep} and in the Inception module \cite{szegedy2015going, szegedy2017inception}.
ResNext \cite{xie2017aggregated} was recently proposed based on modifications of ResNet \cite{he2016deep} where each layer was replaced with a combination of GWC and PWC.
MobileNet \cite{howard2017mobilenets, sandler2018mobilenetv2}, a popular low complexity architecture designed to be implemented in mobile devices,  replaces a  SFCC layer with a DWC followed by a PWC.  For example, consider a $32 \times 32$ input feature map with eight channels that is mapped to sixteen channel output feature map, each also  $32 \times 32$.  With a standard CONV layer, this would take sixteen CONV filters, each $3 \times 3 \times 8$.  In a MobileNet-like layer, the same input and output feature map dimensions would be accomplished first using  eight $3 \times 3 \times 1$ DWC convolutions, followed by sixteen $1 \times 1 \times 8$ point-wise convolutions.  Note that in this example the standard CONV layer has 1,152 parameters and the MobileNet-like layer has only 200 parameters.  
ShuffleNet \cite{zhang2018shufflenet} uses a combination of GWC, a channel shuffling layer, and 
then DWC to provide more information flow across channels. Continuing with the above example, consider a  ShuffleNet-like layer with CONV filters of four groups.  Each group comprises two channels, so that a $3 \times 3 \times 8$ CONV filter is made up of four $3 \times 3 \times 2$ group filters.  Performing the associated GWC yields four output channels.  Using four such CONV filters yields 16 channels, which are each convolved with a different $3 \times 3 \times 1$ filter (i.e., DWC) to produce the final $32 \times 32 \times 16$ OFM.  Thus it would have only 432 parameters.  

Some previous research has indicated that the two-stage processing shown in Fig. \ref{fig:latency_0_1}(a)-(b) may add latency to the computation \cite{singh2019hetconv}, which could impact the speed-up achieved in practice.  This, however, is highly dependent on the implementation.  For example, in an implementation with a multi-core processor (e.g., GPU), the PWC in Fig 2(b) cannot proceed until the GWC is completed. However, in a custom hardware implementation, careful optimization may largely alleviate the performance cost of this bottleneck. Our proposed pre-defined sparse approach, illustrated in Fig. \ref{fig:latency_0_1}(c) and detailed in the next section, avoids this drawback as well. 

\section{Pre-defined Sparse Convolutional Kernels}
\label{sec:spCNN}

We now describe pSConv, our proposed approach to reduce the number of parameters in convolutional neural networks via pre-defined sparse kernels. In pSConv, we exploit the structural sparsity of the input receptive-field (RF) \cite{brendel2019approximating} and propose pre-defined sparse 2D kernel based CONVs to form the channels of each convolutional layer.


\subsection{Pre-Defined Sparse CONV Filters}
\label{sub:psconv}
\begin{figure}[h!]
\includegraphics[width=0.7\linewidth]{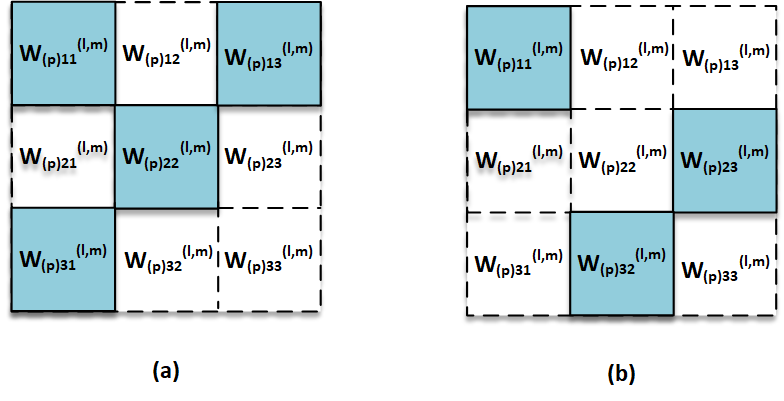}
\centering
\caption{Kernel structure of pre-defined sparse 2D kernels with KS of (a) 4, and (b) 3, respectively. Here $W_{(p)ij}^{(l,m)}$ signifies the $(i,j)^{th}$ element of $p^{th}$ 2D kernel of $m^{th}$ filter in layer $l$.}
\label{fig:pSConv_kernel}
\end{figure}

Assume a convolutional layer $L$ with $C$ CONV filters where each filter is of dimension $k \times k \times C_i^L$ (here, $H_{f}^L$ = $W_{f}^L$ = $k$).\footnote{Typically, $k\in\{1, 3, 5, 7\}$ in modern convolutional neural networks.}  We define a $k \times k \times C_i^L$ pre-defined sparse CONV filter as  one in which some of the $k^2 C_i^L$ elements are fixed to be zero and this pattern is fixed before training and held fixed throughout training and inference. A {\em regular} pre-defined sparse CONV filter has the same kernel support size  for each 2D kernel that comprises the CONV filter. Here, the {\em kernel support size} (KSS), $y\in \{1,\ldots,k^2\}$, is defined to be the number of non-zero weight entries in each $k\times k$ 2D kernel of the CONV filter.\footnote{The kernel support is the set of indices where the kernel is not constrained to be zero.} For example, Fig. \ref{fig:pSConv_kernel}(a) and (b) illustrate instances of kernels with KSS of 4 and 3, respectively. 
The specific method for designing the patterns of kernel support (i.e., constrained zero element patterns) in the CONV filter may vary with implications discussed briefly below.

\begin{figure}[h!]
\includegraphics[width=0.9\linewidth]{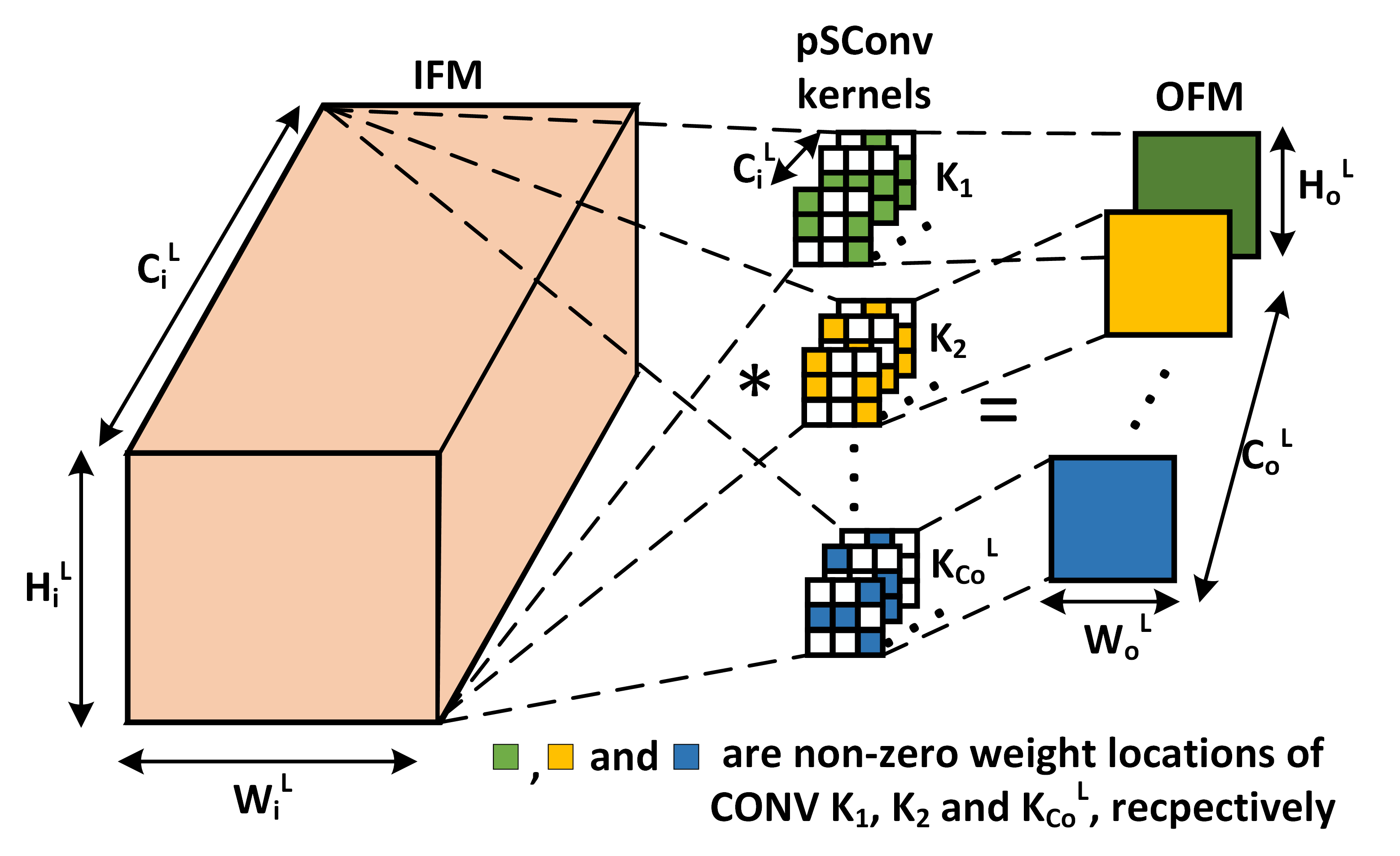}
\centering
\caption{pSConv: Proposed pre-defined sparse kernel based convolution with KSS of 4.}
\label{fig:pSConv}
\end{figure}

For example, for $k=3$, there are $9$ possible values of $y$, with $y=9$ denoting the standard 2D kernel without any pre-defined sparsity, and $y=4$ denotes that only four non-zero entries are allowed in the $3 \times 3$ kernel space while the remaining five entries are $0$. Furthermore, for a given KSS, the pattern of non-zero entries is \textit{pre-defined} at the start of the training process. All results presented in this paper were generated using regular pre-defined sparse CONV filters wherein the kernel support (i.e., patterns of defining the sparsity)  were selected in a constrained, pseudo-random manner. The constraint  enforced  ensured that at least one of the $C_i^L$ 2D kernels has non-zero element for each of the $k^2$ locations. Thus, this procedure provides patterns for sparse CONV layers with significantly fewer parameters,  while the receptive field of the CONV filter remains $k \times k$.
The pSConv concept is illustrated in  Fig. \ref{fig:pSConv} where $C$  regular pre-defined sparse CONV filters using a KSS of 4 (i.e., 5 zero locations in each $3\times 3$ kernel) map an IFM of dimension $H_i^L \times W_i^L \times C_i^L$ to an OFM of dimension $H_o^L \times W_o^L \times C_o^L$.  

\subsection{Training for pSConv}

Pre-defined sparsity in CONV filters has the potential to reduce storage and computation complexity, both during training and inference. Storage may be reduced since only the potentially nonzero elements need to be stored, for instance, with a KSS of 3 for a $3\times 3$ 2D kernel, only $(1/3)^{rd}$ of the weights need to be stored. During the forward processing, only the KSS weights need to be multiplied with the input feature maps. Furthermore, during back-propagation of training, the gradient flow only needs to pass through the kernel support. Realizing these potential complexity reductions in practice, however, requires careful design of the kernel support patterns and the associated memory and computational resources. In software, for example, the storage and computation demands for linked-list storage and the associated array referencing may eliminate a significant fraction of these advantages  associated with sparsity. Similar issues will arise in custom hardware acceleration. However these challenges can be alleviated by designing kernel support patterns algorithmically with a small complexity overhead. 
For example, a very similar problem was solved in \cite{dey2019pre} where it was also shown that these structured connection patterns typically outperformed randomly generated patterns.  Another approach is to utilize more generic sparse tensor product accelerators that may be available in hardware or software libraries \cite{han2016eie}. However, this paper focuses on assessing the potential efficacy of pSConv layers and therefore we do not address the above issues.   Specifically, our implementations targeted rapid software development for experimentation and did not attempt to optimize the complexity for sparsity. 

Our specific implementation used Pytorch as development package and we zeroed out the weights in the locations outside of kernel support  at the start of training and at the end of each mini-batch update -- i.e., the  weights were initially updated without a sparsity assumption and then those outside the kernel support we reset to zero before the next mini-batch. 





\subsection{FLOPs Count}

The FLOPs count for various CONV filter types are shown in Table \ref{tab:flops}.  This assumes a layer with IFM size of $H_i^L \times W_i^L \times C_i^L$ and OFM size $H_o^L \times W_o^L \times C_o^L$.   This is the  number of FLOPs required to perform the forward (i.e. inference) processing.  These expressions assume an efficient implementation as discussed above and are therefore ideal.  Specifically, the overhead with generating address and permutations -- i.e., in ShuffleNet and pSConv -- are not included in these expressions.  Note the the reduction in FLOPs for pSConv relative to the standard SFCC layer is  simply the kernel density -- \i.e. the KSS divided by the size of the standard 2D kernel.

\begin{table}[t]
  \centering
  \begin{tabular}{|c|c|}
  \hline
    Approach &  FLOP Count (Forward, Ideal) \\\hline \hline
    SFCC       & $H_o^L \times W_o^L \times C_i^L \times C_o^L \times k\times k$  \\\hline
    pSConv        & $ H_o^L \times W_o^L \times C_i^L \times C_o^L \times y $  \\\hline
    MobilNet-like     &   $(H_o^L \times W_o^L \times  C_i^L \times k \times k)$ \\
    (DWC-PWC) &  $+ (H_o^L \times W_o^L \times C_i^L \times C_o^L)$ \\\hline
    ShuffleNet-like    &  $(H_o^L \times W_o^L \times   C_i^L \times C_o^L \times k \times k)/G$  \\
    (GWC-DWC) & $+ (H_o^L \times W_o^L \times C_o^L \times k \times k)$ \\\hline
    \end{tabular}
  \caption{Expressions for the  FLOP count for the inference operations for various pre-defined constrained filter designs.}
  \label{tab:flops}
\end{table}


\section{Experimental Evaluations} \label{Experment_eval}

In this section we first give an overview of the experimental settings, and then discuss the results in detail for each experiment.
\subsection{Datasets, Architectures, and Hyperparameters}
We use CIFAR-10 and Tiny ImageNet, two popular image classification datasets, for our evaluation. 
Both of these datasets comprise three-channel colored images.   CIFAR-10 has $10$ distinct classes, and each image  has a size of $(H_i^0,W_i^0,C_i^0) = (32, 32, 3)$, where $H_i^0$,$W_i^0$, and $C_i^0$ denote the input height, width, and number of channels, respectively. Tiny ImageNet has $200$ classes, and each image has a size of $(H_i^0,W_i^0,C_i^0) = (64,64,3)$.

\begin{figure}[h!]
\includegraphics[width=0.8\linewidth]{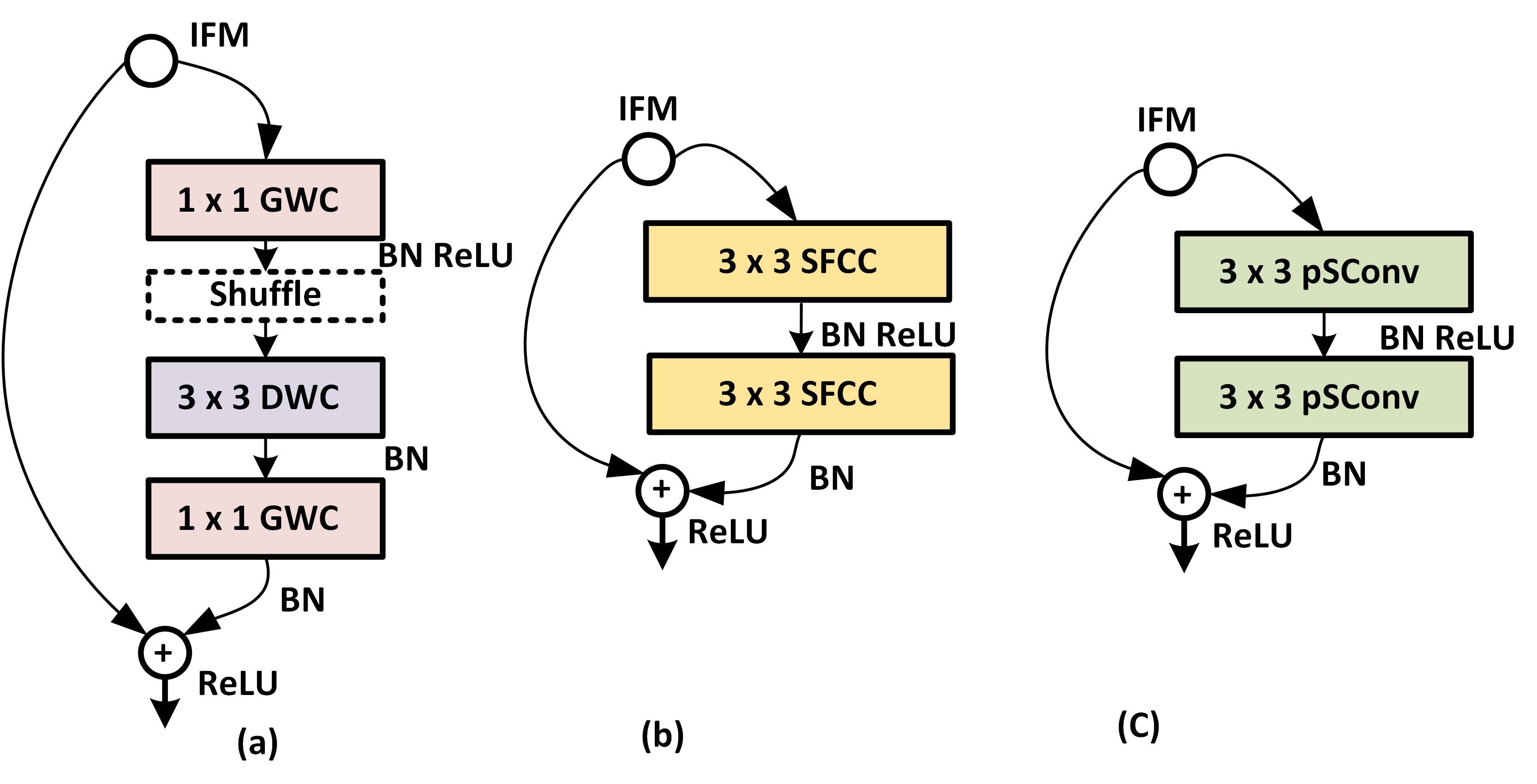}
\centering
\caption{A unit of (a) ShuffleNet bottleneck, (b) ResNet18 basic block, and (c) Modified ResNet18 with pSConv based basic block. }
\label{fig:Shuffle_vs_pSConv}
\end{figure}

We evaluate the benefits of pSConv on two state of the art neural network architectures -- ResNet18 \cite{he2016deep} and VGG16 \cite{simonyan2014very}. We defer the details of the architectures until the associated subsections that follow. For both ResNet18 and VGG16, we modified the level of sparsity in all of the $3 \times 3$ 2D kernels using the constrained, pseudo-random method described in Section \ref{sec:spCNN}, while the fully-connected and down-sampling layers are not modified. 

In each experiment, the number of training epochs is $60$. We use an initial learning rate of $0.1$ with a step decay of $0.2$ after $40^{th}$ and $55^{th}$ epoch. We use the stochastic gradient descent optimizer with a momentum of $0.9$ and weight decay of $0.0005$ for all experiments. A training dataset size of 40,000 with 10,000 each for validation and test is used for CIFAR-10 and training batch size is $128$. For Tiny ImageNet we use a training dataset of 100,000 images and 5000 images each for validation and test. The batch size for training is set to $100$. For each experiment, the final results have been presented for a single training experiment. In the following tables and figures, we denote the networks explored by the base network name, followed by the kernel support size  -- i.e., $\text{<basename>}\_\text{pSC<KSS>}$.  For example, ResNet$18\_\text{pSC4}$ denotes the ResNet18 architecture with $4$ non-zeros ($5$ zeros) in each 2D kernel.

\subsection{Experiments with ResNet18}
\begin{table*}[t]
  \centering
  \begin{tabular}{|c|c|c|c|c|c|}
  \hline
    Model & Test Acc $(\%)$ & FLOPs & FLOPs Reduced (\%) & Parameters & Parameters Reduced (\%) \\\hline
    ResNet18$\_$pSC9       & 91.3 & 0.559 G & ---& 11.17 M& --- \\\hline
    ResNet18$\_$pSC4       & \textbf{91.5} & 0.227 G & 59.4& 5.07 M&  54.6\\\hline
    ResNet18$\_$pSC2       & 90.8 & 0.117 G & 79.1& 2.63 M&  76.5\\\hline
    ResNet18$\_$HC$\_$pSC9 & 89.8 & 0.136 G & 75.7 & 2.8 M&  74.9\\\hline
    ResNet18$\_$HC$\_$pSC4 & 89.3 & 0.056 G & 89.9 & 1.27 M&  88.6\\\hline
    ResNet18$\_$HC$\_$pSC2 & 88.3 & \textbf{0.030 G} & \textbf{94.6}& \textbf{0.66 M} &  \textbf{94.1}\\\hline
    ShuffleNet          & 85.0 & 0.098 G & 82.5& 2.16 M&  80.7\\\hline
    \end{tabular}
  \caption{Performance evaluation of different variants of ResNet18 on CIFAR-10. The standard ResNet18 architecture is used as the reference for calculating the reduction in parameter and FLOP counts. In general, pSConv combined with the standard architectures can give significant improvements in FLOP and parameter counts with negligible drops in accuracy.}
  \label{tab:1}
\end{table*}

\begin{table*}[t]
  \centering
  \begin{tabular}{|c|c|c|c|c|c|}
  \hline
    Model &  Test Acc $(\%)$ & FLOPs & FLOPs Reduced (\%) & Parameters & Parameters Reduced (\%) \\\hline
    ResNet18$\_$pSC9   & \textbf{61.3} & 2.22 G & ---  &11.58 M & --- \\\hline
    ResNet18$\_$pSC4   & 60.7 & 0.902 G & 59.4 &5.47 M & 52.8 \\\hline
    ResNet18$\_$pSC2       & 59.4 & 0.463 G & 79.1 & 3.03 M& 73.8 \\\hline
    ResNet18$\_$HC$\_$pSC9 & 58.5 & 0.561 G & 74.7     & 3.00 M& 74.1 \\\hline
    ResNet18$\_$HC$\_$pSC4 & 57.3 & 0.228 G & 89.7     & 1.47 M& 87.3 \\\hline
    ResNet18$\_$HC$\_$pSC2 & 54.5 & \textbf{0.117 G} & \textbf{94.7}     & \textbf{0.863 M} & \textbf{92.6} \\\hline
    ShuffleNet          & 54.3 & 0.390 G & 82.4     & 3.37 M& 70.9 \\\hline
    \end{tabular}
  \caption{Performance evaluation of different variants of ResNet18 on Tiny ImageNet. The standard ResNet18 architecture is used as the reference for calculating the reduction in parameter and FLOP counts. In general, pSConv combined with the standard architectures   significant reduces the FLOP and parameter counts with negligible degradation in accuracy.}
  \label{tab:2}
\end{table*}

In ResNet18, the first layer is a convolutional layer followed by batch normalization and ReLU. This is followed by $4$ layers each consisting of $2$ basic blocks. As illustrated in Fig. \ref{fig:Shuffle_vs_pSConv}(b), one basic block comprises  two paths for the input. The first path goes through a convolutional layer with $3\times 3$ kernel, followed by batch normalization and ReLU, which is followed by another convolutional layer with $3\times 3$ kernel and batch normalization. The second path is a skip connection, which is directly added to the result from the first path, and then passed through a ReLU. The skip connection path corresponding to the basic block which changes the input feature map dimension consists of $1 \times 1$ convolution layer with stride of 2 to match the dimension before adding. The output of the final basic block is flattened and fed into a fully connected layer with softmax, which has 10 output neurons for CIFAR-10 and $200$ output neurons for Tiny ImageNet. 

\begin{figure}[h!]
\includegraphics[width=0.8\linewidth]{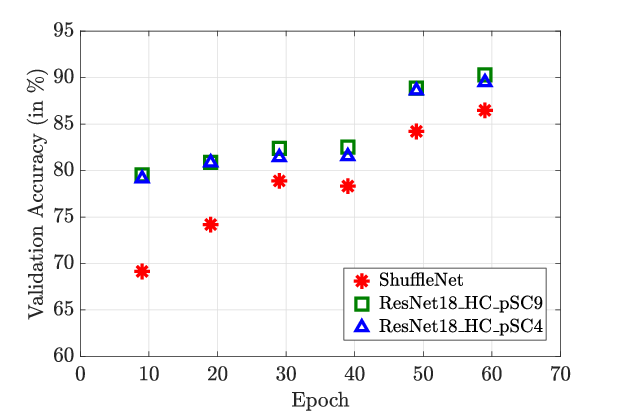}
\centering
\caption{Validation accuracy vs epoch on CIFAR-10. The architectures being compared here are ShuffleNet, ResNet18$\_$HC$\_$pSC9 (standard ResNet18$\_$HC), and ResNet18$\_$HC$\_$pSC4 (ResNet18$\_$HC with 5 zeros (out of 9 elements) in each 2D kernel). }
\label{fig:fig1}
\end{figure}

\begin{figure}[!ht]
\includegraphics[width=0.8\linewidth]{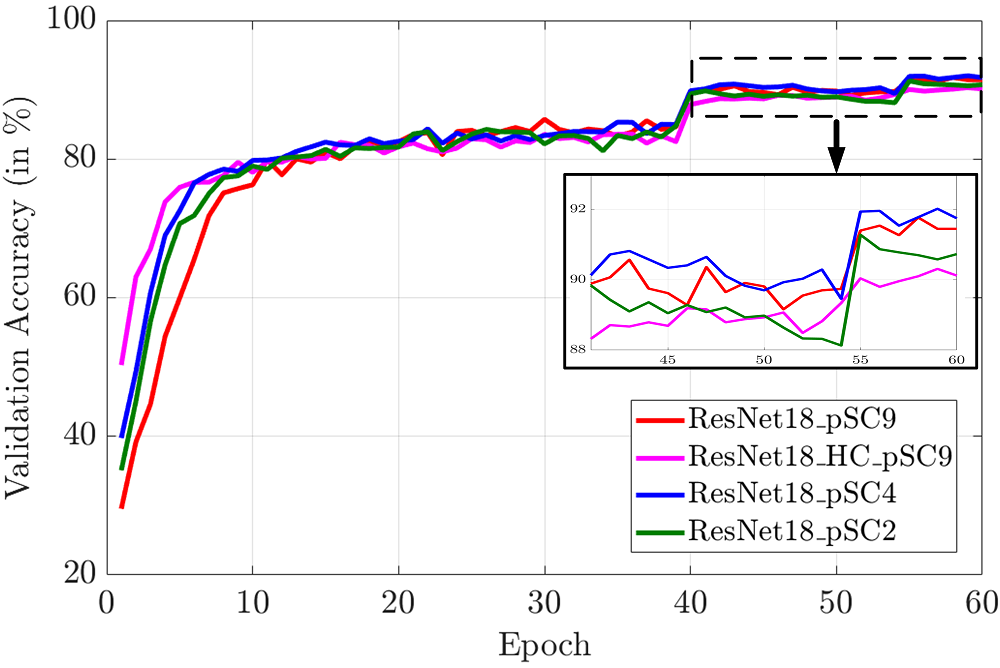}
\vspace*{-0.15in}
\centering
\newline
\caption{Validation accuracy vs epoch on CIFAR-10 for ResNet18$\_$pSC9 (standard architecture), ResNet18$\_$pSC4 (pSConv with $4$ non-zeros in each 2D kernel),  and ResNet18$\_$pSC2 (pSConv with $2$ non-zeros in each 2D kernel).}
\label{fig:fig2}
\end{figure}

 For standard ResNet18 the number of output channels in the first computational layer of 2 basic blocks is 64 and increases to 128, 256, and 512, for the second, third, and fourth computational layers, respectively. In order to more closely match the parameter count with ShuffleNet, we also use a modified version of ResNet18 with a channel width multiplier value of $0.5$. So, this modified version of ResNet18 uses half the number of channels in each of these computational layers (i.e., 32, 64, 128, 256). We refer to this modified  ResNet18 as ``half-channel'' ResNet18 -- i.e., ResNet18$\_$HC. For our work, we use a variant of ShuffleNet with group-size of 3 and channel depths of the stages as 384, 768 and 1536 which has comparable parameters count as ResNet18$\_$HC$\_$pSC9.

We first present our results on CIFAR-10. In Fig. \ref{fig:fig1}, we plot the validation accuracy vs epoch for ShuffleNet and ResNet18$\_$HC combined with pSConv. It shows both the ResNet18$\_$HC variants perform better at all the checkpoint epochs.  Also, it is clear from Table \ref{tab:1} that although ResNet18$\_$HC$\_$pSC4 (with only 4 non-zeros in each 2D kernel) and  ResNet18$\_$HC$\_$pSC2 (with only 2 non-zeros in each 2D kernel) have fewer parameters and FLOPs, they have significantly better test accuracy (89.3$\%$ and 88.3 $\%$, respectively) compared to ShuffleNet ($85.0\%$). 

In Fig. \ref{fig:fig2}, we plot validation accuracy vs epoch for the standard ResNet18, ResNet18$\_$HC and ResNet18 combined with pSConv. 
We see similar trend in validation accuracy improvement (including the jumps at epoch 40 and 55) for all the variants.

\begin{figure}[h!]
\includegraphics[width=0.8\linewidth]{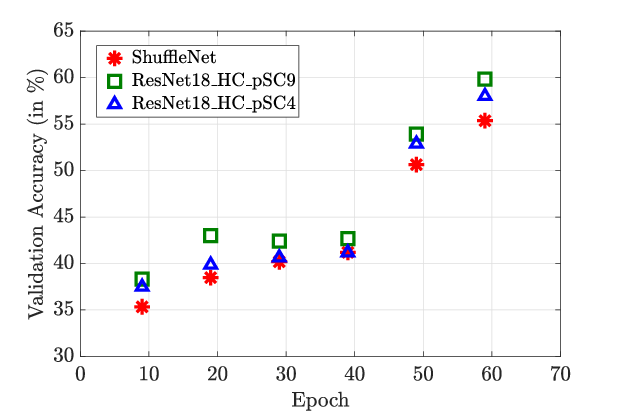}
\centering
\caption{Validation accuracy vs epoch on Tiny ImageNet for ResNet18$\_$HC$\_$pSC9 (standard half-channel ResNet18), ResNet18$\_$HC$\_$pSC4 (pSConv with $4$ non-zeros in each 2D kernel),  and ResNet18$\_$HC$\_$pSC2 (pSConv with $2$ non-zeros in each 2D kernel).}
\label{fig:fig4}
\end{figure}

\begin{figure}[h!]
\includegraphics[width=0.8\linewidth]{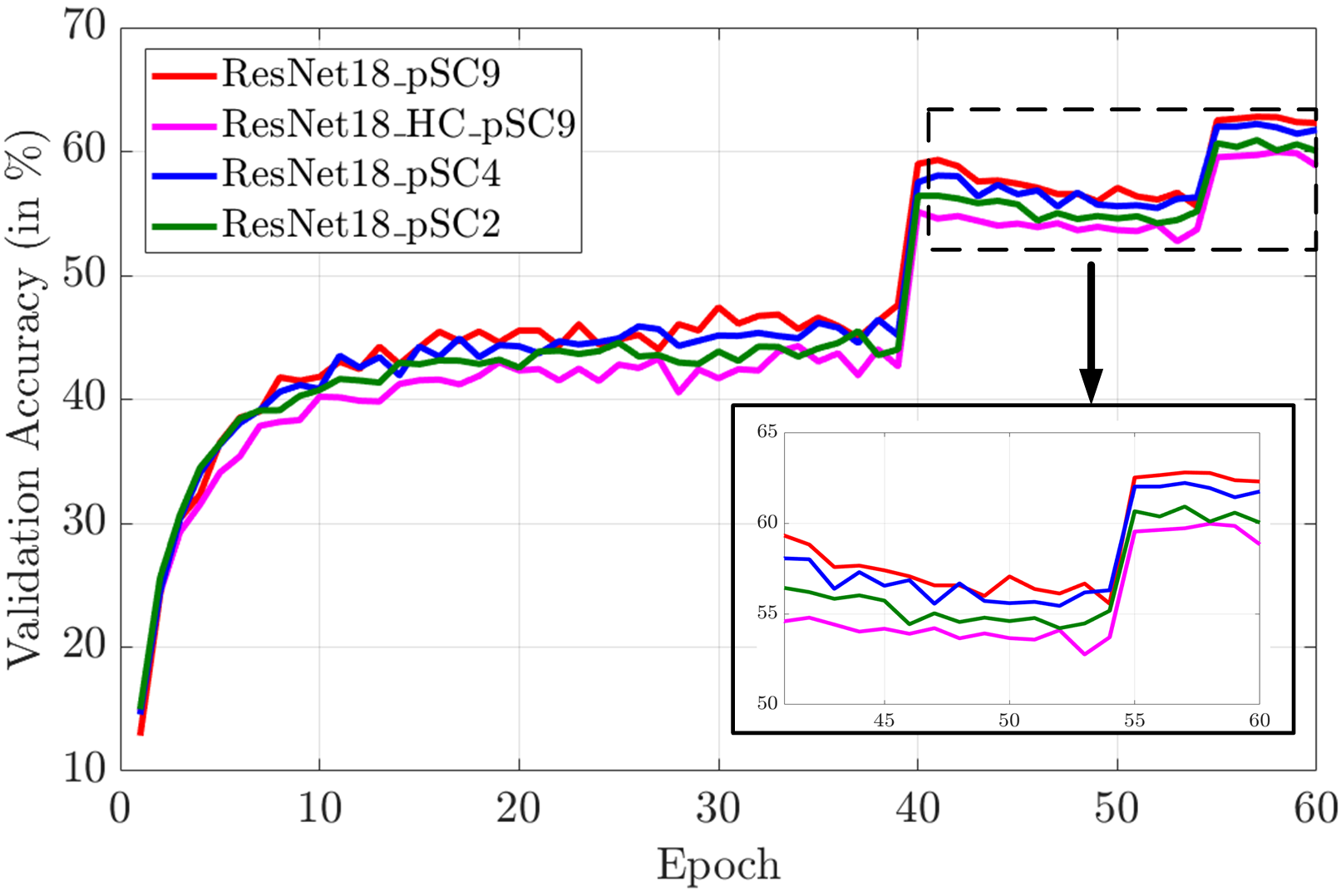}
\vspace*{-0.15in}
\centering
\newline
\caption{Validation accuracy vs epoch on Tiny ImageNet for ResNet18$\_$pSC9 (standard architecture), ResNet18$\_$pSC4 (pSConv with $4$ non-zeros in each 2D kernel),  and ResNet18$\_$pSC2 (pSConv with $2$ non-zeros in each 2D kernel).}
\label{fig:fig5}
\end{figure}

Next we discuss our experimental results with Tiny ImageNet. In Fig. \ref{fig:fig4}, we plot the validation accuracy vs epoch for ShuffleNet and ResNet18$\_$HC combined with pSConv. Table \ref{tab:2} shows the comparison of FLOPs, parameter count and test accuracy. It is noteworthy that  ResNet18$\_$HC$\_$pSC2 (with only 2 non-zeros in each Conv2D kernel) with 3.3$\times$ lesser FLOP count provides similar accuracy as ShuffleNet.

Fig. \ref{fig:fig5} illustrates the validation accuracy vs epoch for the standard ResNet18, ResNet18$\_$HC and ResNet18 combined with pSConv. Clearly, ResNet18$\_$pSC2 and  ResNet18$\_$pSC4 have similar validation accuracy improvement trend as the standard ResNet18, ResNet18$\_$HC. Furthermore, as illustrated in Table \ref{tab:2}, even for larger datasets pSConv can reduce the parameter and FLOP counts of the standard architectures without significant drop in test accuracy. 

\begin{figure}[h!]
\includegraphics[width=0.8\linewidth]{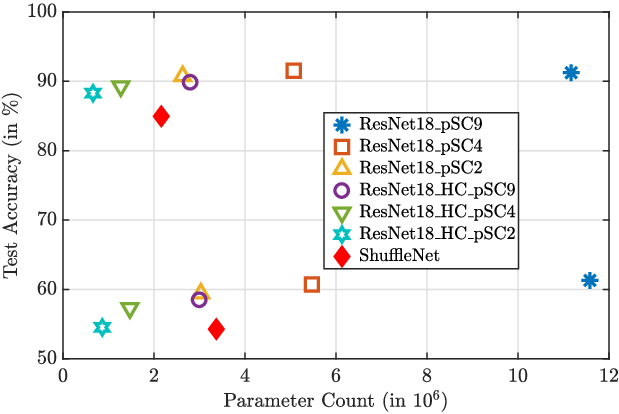}
\centering
\newline
\caption{Test accuracy vs parameter count for the experiments based on ResNet architectures. The results along the top correspond to CIFAR-10, while the results along the bottom correspond to Tiny ImageNet.}
\label{fig:fig11}
\end{figure}

Finally, in Fig. \ref{fig:fig11}, we plot the test accuracy as a function of parameter count for all the experiments related to ResNet18. For both CIFAR-10 and Tiny ImageNet, ResNet18$\_$HC$\_$pSC2 (half-channel ResNet18 with KSS of 2) has the lowest parameter count. With a 3.27$\times$ and 3.9$\times$ reduced parameter count compared to ShuffleNet, ResNet18$\_$HC$\_$pSC2 provides 3.3\% improved and similar accuracy on CIFAR-10 and Tiny ImageNet dataset, respectively. Furthermore, the parameters and FLOPs for ResNet18$\_$HC$\_$pSC9 and ResNet18$\_$pSC2 are quite similar, whereas the latter has around 1\% improved test accuracy for both the datasets.  This is somewhat in agreement with the trends observed in \cite{dey2019pre} for sparse multi-layer perceptrons (MLPs) where it was observed that using more neurons that are sparsely connected typically yields better performance than fewer neurons with full connectivity. Also somewhat surprisingly, the half-channel version of ResNet18 is an attractive, and simple, design alternative to ShuffleNet.


\subsection{Experiments with VGG16}

\begin{table*}[t]
  \centering
  \begin{tabular}{|c|c|c|c|c|c|}
  \hline
    Model & Test Acc $(\%)$ & FLOPs & FLOPs Reduced (\%) & Parameters & Parameters Reduced (\%) \\\hline
    VGG16$\_$pSC9       & 90.2 & 0.333 G & --- &33.65 M  & --- \\\hline
    VGG16$\_$pSC4       & 90.0 & 0.145 G & 56.5 &25.47 M & 24.3 \\\hline
    VGG16$\_$pSC2       & 88.8 & 0.082 G & 75.4 &22.20 M & 34.0 \\\hline
    \end{tabular}
  \caption{Performance evaluation of different variants of VGG16 on CIFAR-10. Applying pSConv results in a parameter reduction of up to $34.0\%$ and FLOP count reduction by up to $75.4 \%$.}
  \label{tab:3}
\end{table*}


In VGG16, there are $13$ convolutional layers, where each convolutional layer is followed by batch normalization and ReLU. There are max pooling layers after ReLU corresponding to the convolutional layers $2$, $4$, $7$, $10$ and $13$. The last max pooling layer feeds an average pooling layer. The next two layers are fully connected layers with ReLUs, while the final layer is a fully connected layer with softmax having 10 output neurons for CIFAR-10 and $200$ output neurons for Tiny ImageNet. 

\begin{figure}[!ht]
\includegraphics[width=0.8\linewidth]{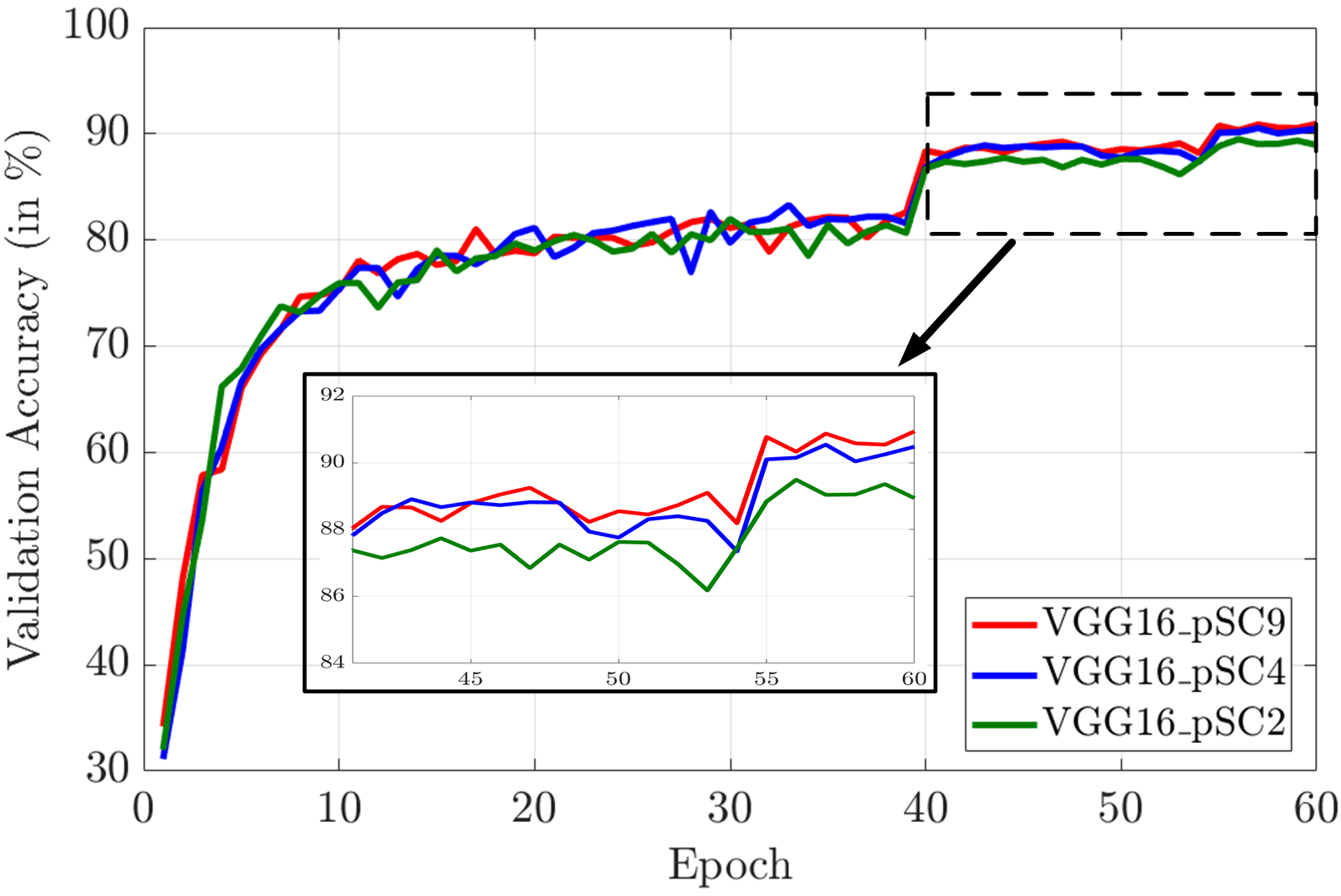}
\vspace*{-0.15in}
\centering
\newline
\caption{Validation accuracy vs epoch on CIFAR-10 for VGG16$\_$pSC9 (standard architecture), VGG16$\_$pSC4 (pSConv with $4$ non-zeros in each 2D kernel),  and VGG16$\_$pSC2 (pSConv with $2$ non-zeros in each 2D kernel).}
\label{fig:fig8}
\end{figure}

For VGG16 with CIFAR-10, we summarize the results in Table \ref{tab:3}, where we can clearly see that with a decrease in KSS, both parameter and FLOPs decrease significantly, with a modest decrease in test accuracy. As illustrated by Fig.\ref{fig:fig8}, architectures with pSConv have a similar convergence curve for validation accuracy as that of the standard architecture. 

\begin{table*}[t]
  \centering
  \begin{tabular}{|c|c|c|c|c|c|}
  \hline
    Model & Test Acc $(\%)$ & FLOPs & FLOPs Reduced (\%) & Parameters & Parameters Reduced (\%) \\\hline
    VGG16$\_$pSC9       &  54.2 &  1.28 G & --- & 40.72 M  & --- \\\hline
    VGG16$\_$pSC4       &  53.5 &  0.528 G & 58.8 & 32.54 M & 20.1 \\\hline
    VGG16$\_$pSC2       &  51.8 &  0.277 G & 78.4 & 29.28 M & 28.1 \\\hline
    \end{tabular}
  \caption{Performance evaluation of different variants of VGG16 on Tiny ImageNet. Applying pSConv results in a parameter reduction of up to $28.1\%$ and FLOP count reduction by up to $78.4 \%$.}
  \label{tab:4}
\end{table*}

\begin{figure}[!ht]
\includegraphics[width=0.8\linewidth]{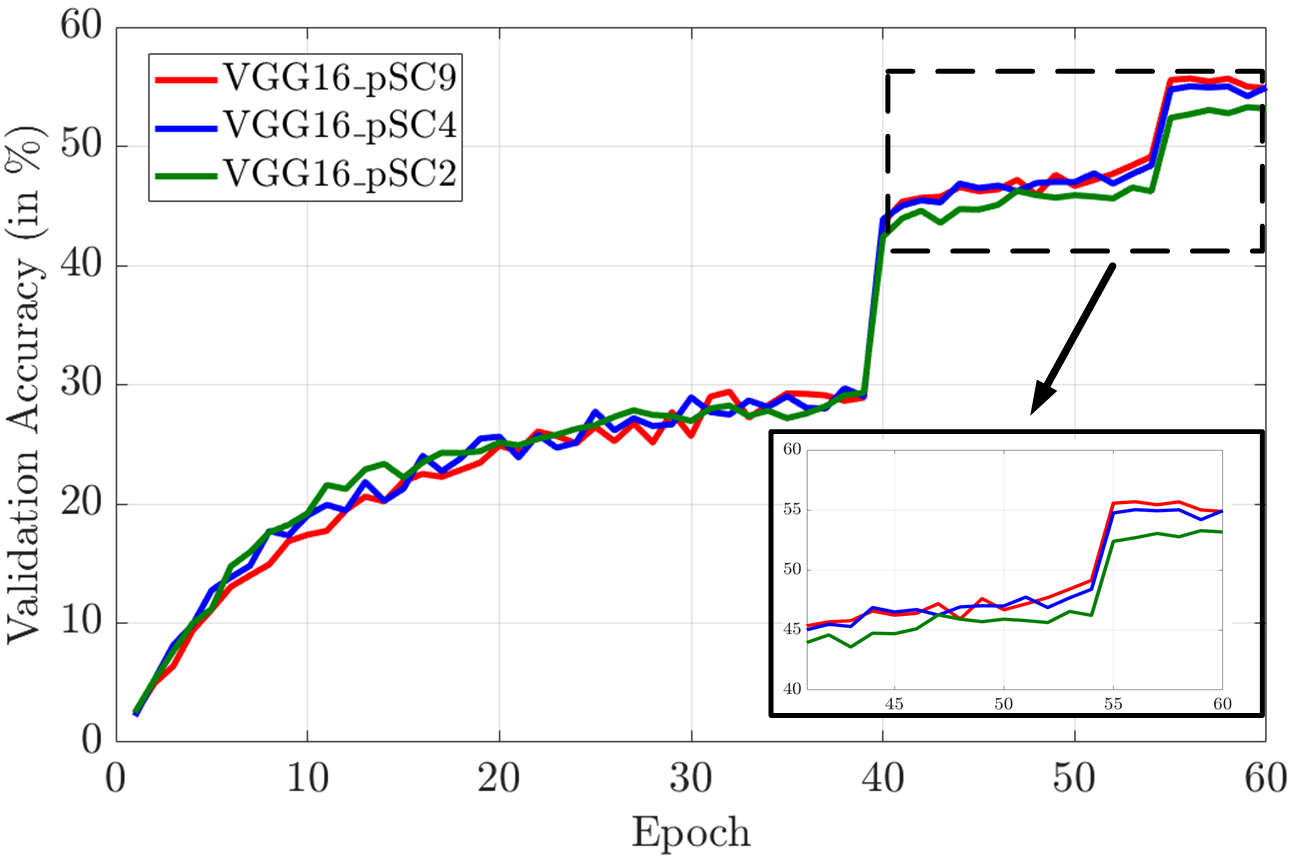}
\vspace*{-0.15in}
\centering
\newline
\caption{Validation accuracy vs epoch on Tiny ImageNet for VGG16$\_$pSC9 (standard architecture), VGG16$\_$pSC4 (pSConv with $4$ non-zeros in each 2D kernel),  and VGG16$\_$pSC2 (pSConv with $2$ non-zeros in each 2D kernel).}
\label{fig:fig9}
\end{figure}

Similar results are obtained for training over Tiny ImageNet. As shown in Table \ref{tab:4}, as KSS is decreased, both parameter and FLOPs count decrease significantly, with only a modest decrease in test accuracy. Specifically, we see a reduction of parameter and FLOPs count by $28.1\%$ and $78.4\%$, respectively.\footnote{VGG has a larger portion of its parameters  in the fully-connected classification layers than ResNet18, so the FLOPs reduction and parameter count reductions differ less for VGG than for ResNet18.} Furthermore, as illustrated by Fig. \ref{fig:fig9}, architectures with pSConv have a similar convergence curve for validation accuracy as that of the standard architecture.

\section{Conclusions and future work}
\label{sec:concl}
The proposed pre-defined sparse kernel based filter design approach (pSConv) can achieve reduced complexity for training and inference while yielding higher accuracy than start-of-the-art resource-constrained alternatives. In our approach, only a subset of kernel weights, referred to as the kernel support, are not fixed at zero. 
Our evaluations with CIFAR-10 dataset have shown a ResNet18 with half the channel size at each layer and with a kernel support size  of 2 out performs ShuffleNet by an absolute accuracy margin of around $3.3\%$ with a $3.27 \times$ reduction in the number of parameters.   Similar trends were observed for Tiny ImageNet dataset where a similar ResNet18 architecture with a kernel support size 4 provides  $~3\%$ higher accuracy with $ 2.29 \times$ reduction in the number of parameters.  

While the results of this work demonstrate the potential of pre-defined sparsity in CNNs, there are several interesting areas for further research.  First, since pre-defined sparsity and group-wise/point-wise convolutions are orthogonal methods, a more complete network architecture search that optimizes over these pre-defined constrained filter methods could be fruitful.  Second, investigating efficient implementations of pSConv layers for software (GPU) and custom hardware implementations is an important area for future work.  
\section{Acknowledgements}
\label{sec:ackn}
This paper is an extension of work done by the first three authors in the Spring 2019 offering of EE599: Deep Learning at University of Southern California, taught by Dr.~Keith Chugg.  We would like to thank Dr.~Brandon Franzke, and Dr.~Massoud Pedram for their helpful feedback on this work. Financial support from the National Science Foundation grant \#1763747 and Amazon Educate program which provided cloud compute credits for the course is gratefully acknowledged.

\bibliographystyle{ieeetr}
\vspace{-2mm}
\bibliography{biblio}

\begin{thebibliography}{10}

\bibitem{he2016deep}
K.~He, X.~Zhang, S.~Ren, and J.~Sun, ``Deep residual learning for image
  recognition,'' in {\em Proceedings of the IEEE Conference on Computer Vision
  and Pattern Recognition}, pp.~770--778, 2016.

\bibitem{girshick2014rich}
R.~Girshick, J.~Donahue, T.~Darrell, and J.~Malik, ``Rich feature hierarchies
  for accurate object detection and semantic segmentation,'' in {\em
  Proceedings of the IEEE conference on computer vision and pattern
  recognition}, pp.~580--587, 2014.

\bibitem{krizhevsky2012imagenet}
A.~Krizhevsky {\em et~al.}, ``{ImageNet} classification with deep convolutional
  neural networks,'' in {\em Advances in Neural Information Processing
  Systems}, pp.~1097--1105, 2012.

\bibitem{young2018recent}
T.~Young, D.~Hazarika, S.~Poria, and E.~Cambria, ``Recent trends in deep
  learning based natural language processing,'' {\em IEEE Computational
  Intelligence Magazine}, vol.~13, no.~3, pp.~55--75, 2018.

\bibitem{abdel2014convolutional}
O.~Abdel-Hamid, A.-r. Mohamed, H.~Jiang, L.~Deng, G.~Penn, and D.~Yu,
  ``Convolutional neural networks for speech recognition,'' {\em IEEE/ACM
  Transactions on Audio, Speech, and Language Processing}, vol.~22, no.~10,
  pp.~1533--1545, 2014.

\bibitem{lecun1998mnist}
Y.~LeCun, ``The {MNIST} database of handwritten digits,'' {\em http://yann.
  lecun. com/exdb/mnist/}, 1998.

\bibitem{coates2013deep}
A.~Coates {\em et~al.}, ``Deep learning with {COTS HPC} systems,'' in {\em
  International Conference on Machine Learning}, pp.~1337--1345, 2013.

\bibitem{hegde2018ucnn}
K.~Hegde, J.~Yu, R.~Agrawal, M.~Yan, M.~Pellauer, and C.~W. Fletcher, ``{UCNN}:
  Exploiting computational reuse in deep neural networks via weight
  repetition,'' in {\em Proceedings of the 45th Annual International Symposium
  on Computer Architecture}, pp.~674--687, IEEE Press, 2018.

\bibitem{courbariaux2015binaryconnect}
M.~Courbariaux, Y.~Bengio, and J.-P. David, ``{BinaryConnect}: Training deep
  neural networks with binary weights during propagations,'' in {\em Advances
  in Neural Information Processing Systems}, pp.~3123--3131, 2015.

\bibitem{zhou2017incremental}
A.~Zhou, A.~Yao, Y.~Guo, L.~Xu, and Y.~Chen, ``Incremental network
  quantization: {Towards} lossless {CNNs} with low-precision weights,'' {\em
  arXiv preprint arXiv:1702.03044}, 2017.

\bibitem{rastegari2016xnor}
M.~Rastegari, V.~Ordonez, J.~Redmon, and A.~Farhadi, ``{XNOR-Net}: {ImageNet}
  classification using binary convolutional neural networks,'' in {\em European
  Conference on Computer Vision}, pp.~525--542, Springer, 2016.

\bibitem{han2015learning}
S.~Han, J.~Pool, J.~Tran, and W.~Dally, ``Learning both weights and connections
  for efficient neural network,'' in {\em Advances in Neural Information
  Processing Systems}, pp.~1135--1143, 2015.

\bibitem{guo2016dynamic}
Y.~Guo, A.~Yao, and Y.~Chen, ``Dynamic network surgery for efficient {DNNs},''
  in {\em Advances in Neural Information Processing Systems}, pp.~1379--1387,
  2016.

\bibitem{mao2017exploring}
H.~Mao, S.~Han, J.~Pool, W.~Li, X.~Liu, Y.~Wang, and W.~J. Dally, ``Exploring
  the regularity of sparse structure in convolutional neural networks,'' {\em
  arXiv preprint arXiv:1705.08922}, 2017.

\bibitem{wen2016learning}
W.~Wen, C.~Wu, Y.~Wang, Y.~Chen, and H.~Li, ``Learning structured sparsity in
  deep neural networks,'' in {\em Advances in Neural Information Processing
  Systems}, pp.~2074--2082, 2016.

\bibitem{dey2019pre}
S.~Dey, K.-W. Huang, P.~A. Beerel, and K.~M. Chugg, ``Pre-defined sparse neural
  networks with hardware acceleration,'' {\em IEEE Journal on Emerging and
  Selected Topics in Circuits and Systems}, 2019.

\bibitem{simonyan2014very}
K.~Simonyan and A.~Zisserman, ``Very deep convolutional networks for
  large-scale image recognition,'' {\em arXiv preprint arXiv:1409.1556}, 2014.

\bibitem{krizhevsky2009learning}
A.~Krizhevsky and G.~Hinton, ``Learning multiple layers of features from tiny
  images,'' tech. rep., Citeseer, 2009.

\bibitem{le2015tiny}
Y.~Le and X.~Yang, ``Tiny {ImageNet} visual recognition challenge,'' {\em CS
  231N}, 2015.

\bibitem{zhang2018shufflenet}
X.~Zhang, X.~Zhou, M.~Lin, and J.~Sun, ``{ShuffleNet}: An extremely efficient
  convolutional neural network for mobile devices,'' in {\em Proceedings of the
  IEEE Conference on Computer Vision and Pattern Recognition}, pp.~6848--6856,
  2018.

\bibitem{howard2017mobilenets}
A.~G. Howard, M.~Zhu, B.~Chen, D.~Kalenichenko, W.~Wang, T.~Weyand,
  M.~Andreetto, and H.~Adam, ``{MobileNets}: Efficient convolutional neural
  networks for mobile vision applications,'' {\em arXiv preprint
  arXiv:1704.04861}, 2017.

\bibitem{lecun1999object}
Y.~LeCun, P.~Haffner, L.~Bottou, and Y.~Bengio, ``Object recognition with
  gradient-based learning,'' in {\em Shape, contour and grouping in computer
  vision}, pp.~319--345, Springer, 1999.

\bibitem{vanhoucke2014learning}
V.~Vanhoucke, ``Learning visual representations at scale,'' {\em ICLR invited
  talk}, vol.~1, p.~2, 2014.

\bibitem{szegedy2015going}
C.~Szegedy, W.~Liu, Y.~Jia, P.~Sermanet, S.~Reed, D.~Anguelov, D.~Erhan,
  V.~Vanhoucke, and A.~Rabinovich, ``Going deeper with convolutions,'' in {\em
  Proceedings of the IEEE Conference on Computer Vision and Pattern
  Recognition}, pp.~1--9, 2015.

\bibitem{ioannou2017deep}
Y.~Ioannou, D.~Robertson, R.~Cipolla, and A.~Criminisi, ``{Deep Roots}:
  Improving {CNN} efficiency with hierarchical filter groups,'' in {\em
  Proceedings of the IEEE Conference on Computer Vision and Pattern
  Recognition}, pp.~1231--1240, 2017.

\bibitem{szegedy2017inception}
C.~Szegedy, S.~Ioffe, V.~Vanhoucke, and A.~A. Alemi, ``Inception-v4,
  inception-{ResNet} and the impact of residual connections on learning,'' in
  {\em Thirty-First AAAI Conference on Artificial Intelligence}, 2017.

\bibitem{xie2017aggregated}
S.~Xie, R.~Girshick, P.~Doll{\'a}r, Z.~Tu, and K.~He, ``Aggregated residual
  transformations for deep neural networks,'' in {\em Proceedings of the IEEE
  Conference on Computer Vision and Pattern Recognition}, pp.~1492--1500, 2017.

\bibitem{sandler2018mobilenetv2}
M.~Sandler, A.~Howard, M.~Zhu, A.~Zhmoginov, and L.-C. Chen, ``{MobileNetV2}:
  Inverted residuals and linear bottlenecks,'' in {\em Proceedings of the IEEE
  Conference on Computer Vision and Pattern Recognition}, pp.~4510--4520, 2018.

\bibitem{singh2019hetconv}
P.~Singh, V.~K. Verma, P.~Rai, and V.~P. Namboodiri, ``{HetConv}:
  {Heterogeneous Kernel-Based Convolutions for Deep CNNs},'' in {\em
  Proceedings of the IEEE Conference on Computer Vision and Pattern
  Recognition}, pp.~4835--4844, 2019.

\bibitem{brendel2019approximating}
W.~Brendel and M.~Bethge, ``{Approximating CNNs with Bag-of-local-Features
  models works surprisingly well on ImageNet},'' {\em ICLR}, 2019.

\bibitem{han2016eie}
S.~Han, X.~Liu, H.~Mao, J.~Pu, A.~Pedram, M.~A. Horowitz, and W.~J. Dally,
  ``{EIE: Efficient} inference engine on compressed deep neural network,'' in
  {\em 2016 ACM/IEEE 43rd Annual International Symposium on Computer
  Architecture (ISCA)}, pp.~243--254, IEEE, 2016.

\end{thebibliography}

\end{document}